%% file: main.tex
\documentclass{article}
\usepackage{spconf,amsmath,graphicx}
\usepackage{enumitem}
\setlist{nosep, leftmargin=14pt}
\usepackage[table]{xcolor} 
\usepackage{mwe} 
\usepackage{color}
\usepackage[numbers]{natbib}
\usepackage[skip=0.4pt]{caption}
\usepackage{subfigure}

\usepackage[table]{xcolor}
\usepackage{multirow}
\usepackage{rotating}   

\title{TrackletGPT\\ :a language-like GPT framework for White Matter Tract Segmentation}

\name{
    \begin{tabular}{c}
    Anoushkrit Goel$^{1}$, Simroop Singh$^{1}$, Ankita Joshi$^{1}$, Ranjeet Ranjan Jha$^{2}$,\\ Chirag Ahuja$^{3}$, Aditya Nigam$^{1}$, Arnav Bhavsar$^{1}$
    \end{tabular}
}
\address{
    $^1$School of Computing and Electrical Engineering, Indian Institute of Technology Mandi \\
    $^2$Department of Mathematics, Indian Institute of Technology Patna \\
    $^3$Post Graduate Institute of Medical Education Research, Chandigarh \\
}

\address{
    $^1$SCEE, IIT Mandi
    $^2$Department of Mathematics, IIT Patna
    $^3$PGIMER, Chandigarh
}

\begin{document}
\maketitle
\begin{abstract}

White Matter Tract Segmentation is imperative for studying brain structural connectivity, neurological disorders and neurosurgery.
This task remains complex, as tracts differ among themselves, across subjects and conditions, yet have similar 3D structure across hemispheres and subjects.
To address these challenges, we propose \textbf{TrackletGPT}, a language-like GPT framework which reintroduces sequential information in tokens using \textbf{tracklets}.
TrackletGPT generalises seamlessly across datasets, is fully automatic, and encodes granular sub-streamline segments, Tracklets, scaling and refining GPT models in Tractography Segmentation.
Based on our experiments, \textbf{TrackletGPT} outperforms state-of-the-art methods on average DICE, Overlap and Overreach scores on TractoInferno and HCP datasets, even on inter-dataset experiments.

\end{abstract}
\begin{keywords}
Diffusion MRI, Tractography, Deep Learning, Point Cloud, GPT, Auto-Regressive models
\end{keywords}
\vspace{-0.3cm}
\section{Introduction} \label{sec:intro}
Streamlines are tractography algorithms' estimates in 3D derived from fODF (fiber orientation density functions) obtained during Brain Diffusion MRI \cite{basser1994mr} which are segmented into White Matter Tracts for practical use. White Matter Tract Segmentations allows for a non-invasive study of the brain and is an active research problem with works ranging from classical to learnable methods. Classical Methods utilize Mean Direct Flip (\textit{MDF}) Distance to cluster streamlines (\textit{QuickBundles}\cite{garyfallidis2012quickbundles}), recognize bundles (\textit{RecoBundles}\cite{garyfallidis2018recognition}), and search similar streamlines (\textit{Fast Streamline Search} \cite{st2022fast}). However, learnable methods either employ CNNs across MRI slices (\textit{TractSeg}\cite{wasserthal2018tractseg}), use novel fiber descriptors with CNNs (\textit{DeepWMA}\cite{zhang2020deep}), filter in embedding space (\textit{FINTA} \cite{legarreta2021filtering} \& \textit{FIESTA} \cite{dumais2023fiesta, legarreta2023generative}), embed multiple representations in different embeddings (\textit{TractoEmbed}\cite{goel2024tractoembed}), uses point clouds as a streamline representation format (\textit{TractCloud}\cite{xue2023tractcloud}), or encode sequence of point patches in GPT framework (\textit{TractoGPT}\cite{goel2025tractogpt}).

Here most methods either require registration, ATLAS, filtering, calibration for thresholds or multiple representations and yet fail to capture maximum spatial information to segment white matter tracts.
In this paper, 
\begin{itemize} 
\item We introduce \textbf{TrackletGPT}, a language-like GPT framework, which generalizes seamlessly across datasets, is fully automatic and registration-free and achieves state-of-the-art performance on respective datasets.
\item We introduce \textbf{Tracklets}, a novel representation of streamlines, which captures sequential and spatial information of the streamlines. We model tracklets as words, streamlines as sentences, and clusters as paragraphs. 
\end{itemize}
\vspace{-0.3cm}
\section{Methodology} \label{sec:method}


\subsection{dMRI Datasets and Tractography}
\label{sec:data}


For the purpose of this study, we use \textbf{TractoInferno}\cite{poulin2022tractoinferno} and \textbf{HCP} \cite{wasserthal2018tractseg} datasets for training and testing TrackletGPT to provide results on inter-dataset experiments and combined dataset. 

\setlength{\arrayrulewidth}{0.1pt}

\begin{table}[htb!]
    \centering
    \begin{tabular}{p{2.2cm}p{2.7cm}p{0.5cm}p{1.3cm}}
    \hline
    \textbf{Public Dataset} & \textbf{Track, Segment} & \textbf{Class} & \textbf{Subjects} \\
    \hline
    \textbf{HCP} \cite{van2013wu, wasserthal2018tractseg} gold-standard &
    MRtrix3 , TractSeg\cite{wasserthal2018tractseg} &
    72 &
    80:15:10 \\
    \hline
    \textbf{TractoInferno} \cite{poulin2022tractoinferno} 
    
    silver-standard&
    Ensemble 4 Tracking Methods\cite{poulin2022tractoinferno}, \t RecobundlesX\cite{garyfallidis2018recognition} &
    32 &
    198:58:28 \\
    \hline
    \end{tabular}
    \caption{\textit{Subjects} column, denotes ratios of subjects in each split \textbf{train:val:test} and Tracking and Segmentation Methods to generate ground truth bundles and streamlines are listed in the respective columns.}
    \label{tab:datasets}
\end{table}


\begin{figure*}[!hbt]
    \centering
    \includegraphics[width=1\linewidth]{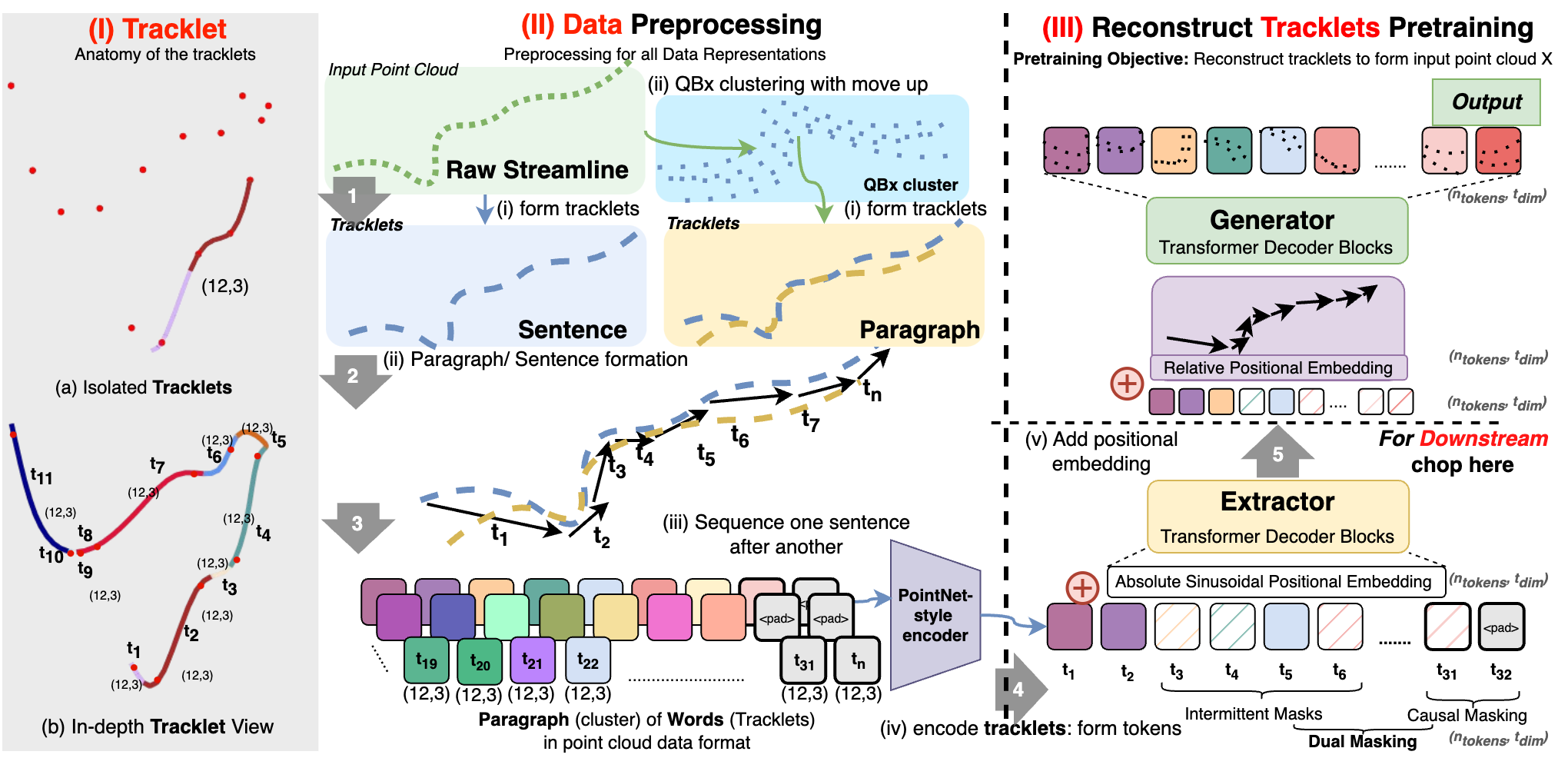}
    \caption{\textbf{TrackletGPT Framework}: (I) Tracklets are formed by connecting inflection points obtained from downsampling interpolation, (II) Raw streamline can either (i) form sentences or (ii) form paragraphs along with neighbouring streamlines. Sequenced sentences(iii) are randomly ordered aggregate to $n$ tracklets where each bicubic-interpolated (12,3) tracklet is (iv) encoded by Pointnet-style encoder to give $(n_{tokens}, t_{dim})$. These tokens along with (v) positional embedding are sent to the GPT framework to reconstruct tracklet tokens.}
    \label{fig:landing}
\end{figure*}

\vspace{-0.3cm}
\subsection{Data Representation} \label{sec:tracklet}
\textbf{Words} \textbf{(Tracklets)} are streamline segments created by dissecting streamlines at inflection points (refer Fig. \ref{fig:landing}).
We tune parameters such that inflection points correspond to the least number of points needed to form a downsampled streamline preserving keypoints and shape.
To embed bi-directional connecting information among tracklets, we extend 2 or more points in each tracklet along the streamline in both directions, this creates overlapping tracklets facilitating connection information among tracklets. 

Tracklets fuse the inherent \textit{sequential} nature of streamlines, with self and neighbouring \textit{spatial} information. 
Tracklets \textit{remove the computational overhead} of finding neighbouring points through FPS-kNN as in TractoGPT\cite{goel2025tractogpt}.

\textbf{Sentences} \textbf{(streamlines)} vary in size, shape, location, and context in the 3D. These expert-annotated streamlines with labels are \textit{input} to TrackletGPT. 
Streamlines are modeled as a \textit{sentence} of bicubic-interpolated tracklets (words) with configurable dimension of (12,3) each, with max words $(n_{tokens}):32$ in a sentence, yielding an array of (32,12,3) dims. Streamlines are padded with (0,0,0) to form a 32-word sentence, which are masked out when fed to the Extractor and Generator (refer Figure \ref{fig:landing}).

\textbf{Paragraphs} \textbf{(clusters)} are formed by \textit{QuickbundlesX clustering with move up}\cite{goel2025tractogpt} where a tree of clusters is initially formed at different resolutions: [40,30,20,10,8,6,4] mm. Among the finer resolution (4, 6, 8mm) clusters, 7 out of 10 neighbouring streamlines (7 neighbours + 1 self) are randomly sampled yielding 8*32 tracklets of (12,3) dims (refer Figure \ref{fig:landing}). Sentences in Paragraphs are randomly ordered in training to learn spatial and not sequential information among each other.

\begin{figure*}[hbt!]
    \centering
    \includegraphics[width=1\textwidth]{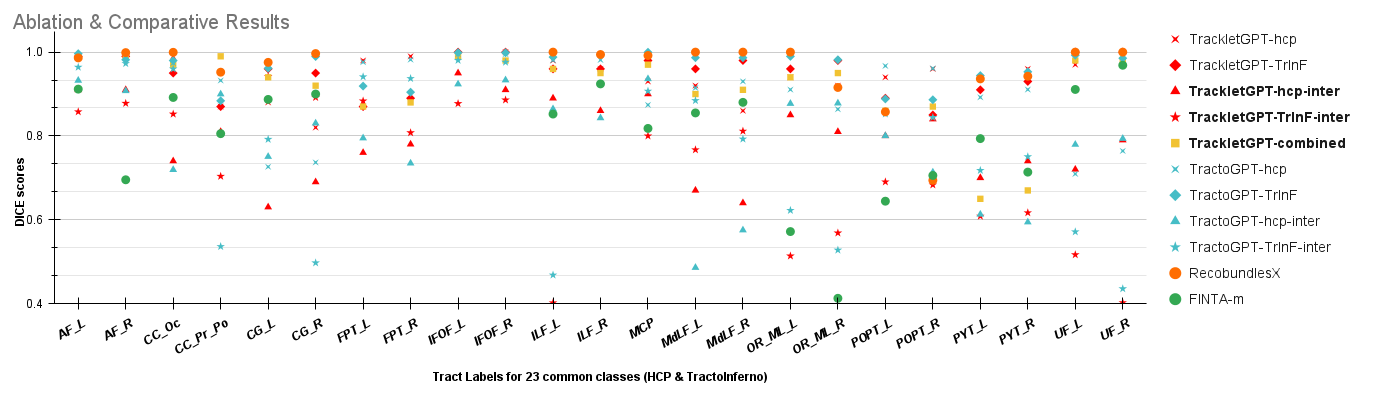}
    \caption{\textbf{Voxel DICE scores for class-wise comparison} across \textit{FINTA-m} \textit{RecoBundlesX}, \textit{TrackletGPT}, \textit{TractoGPT} methods on TractoInferno and HCP, showcasing results on same (self), different (inter-dataset) and combined train test datasets.}
    \label{graph}
\end{figure*}
\vspace{-0.3cm}
\subsection{Model Input}\label{sec:token}

Input tokens for TrackletGPT are generated using a Pointnet-style encoder, which encodes a batch of $n$ tracklets of (12,3) dimensions into a token embedding of dimensions $t_{dim}$, where $t_{dim}$ depends on the size of the model $t_{dim}$: [M:mini (256), S:small (384), B:big (512)]. Based on our experiments, $mini$ model outperforms other variants in most of the cases, keeping TrackletGPT a light 9.8M parameter model. 
Bi-directional sequencing of tracklets (refer Figure \ref{fig:landing})\cite{chen2024pointgpt} along the streamlines are encoded as a 1D array ($t_{dim}$) length for $n_{tokens}$ each.
\vspace{-0.4cm}
\subsection{TrackletGPT Model}
\textbf{TrackletGPT} follows an autoregressive GPT-style framework in which the model is first pretrained to reconstruct tracklets and later adapted for streamline classification. TrackletGPT uses stacked transformer decoder blocks, with an \textit{Extractor} learning latent representations of tracklets and a \textit{Generator} predicting tracklets autoregressively \cite{chen2024pointgpt, yu2022point, goel2025tractogpt}.

Tracklet reconstruction in pretraining uses a dual-masking strategy that combines causal and intermittent random masking, reducing short-range dependencies and mitigating overfitting. The Extractor first obtains point-level embeddings using a PointNet-style encoder and directly adds sinusoidal positional encodings computed from normalized center points of tracklets. 
It is followed by stacked transformer decoder layers whose depth varies by model size: 8 layers for M (Mini, 9.8M params), 12 layers for S (Small, 29.2M params), and 16 layers for B (Big, 69.3M params). The Generator consists of 4 transformer decoder blocks and incorporates relative position embedding (Fig.~\ref{fig:landing}, Stages~II--III) to provide directional structure without revealing masked regions, enabling autoregressive tracklet reconstruction.
For downstream classification, we use the Extractor’s latent representations, attaching a compact classification head of stacked linear--BN--ReLU layers with dropout and finetune it with supervised class labels.


\begin{figure}[hbt!]
    \centering
    \includegraphics[width=1\linewidth]{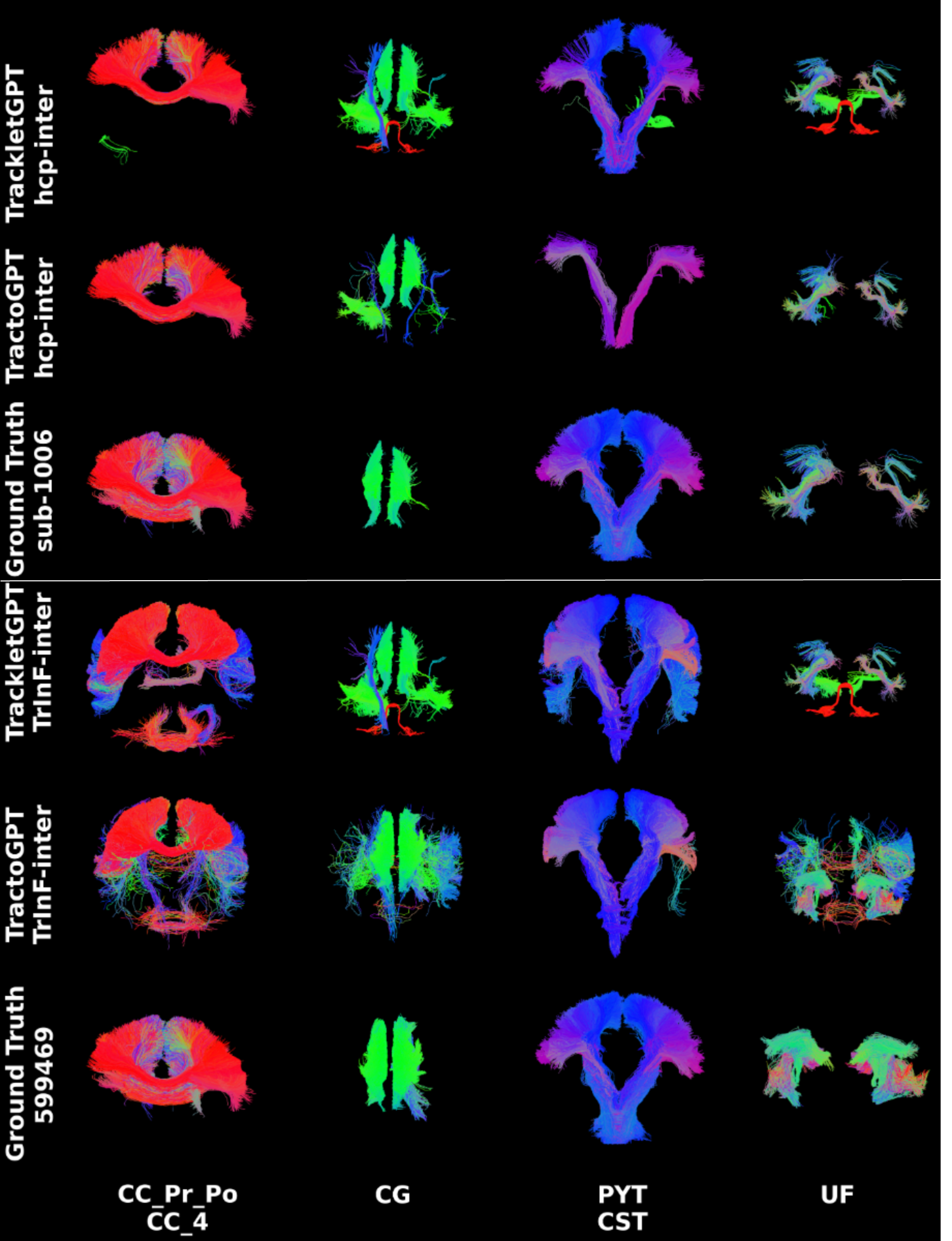}
    \caption{Visualisation of Inter-dataset test results for sub-1006 (Tractoinferno) \& 599469 (HCP)}
    \label{plots}
\end{figure}
\vspace{-0.3cm}
\subsection{Model Training and Testing}
To mitigate class imbalance, we sample up to 1k streamlines per bundle and then select a uniform set of 10k streamlines per subject for training. However at test time, all streamlines of each subject are processed.

\textbf{TrackletGPT} is trained in two stages: self-supervised pretraining followed by supervised fine-tuning. During pretraining, the model learns to reconstruct streamline from sequences of tracklets on Chamfer Distance Loss (equal L1 and L2 components) with AdamW (weight decay 0.05) optimizer and a cosine learning rate schedule initialized at 1e{-4}, which typically converges earlier than the 300-epoch limit.
Supervised classification fine-tuning combines cross entropy loss and Chamfer Distance Loss (CDL1 + CDL2) in 1:3 proportions. This finetuning promotes model to learn both discriminative accuracy and geometric consistency. \textbf{TrackletGPT} on RTX3090 converges under 48hrs, and infers all streamlines in a subject taking 25 mins (Sentence) to 50 mins (Paragraph).

\vspace{-0.4cm}
\section{Experiments and Results}
In this section, we present comparative, ablation and inter-dataset studies to assess TrackletGPT.
We report DICE scores for consistency across methods, in \textbf{Ablation} Study (refer Table \ref{tab:ablation}) we see Paragraph representation performs better due to the neighbouring information. In \textbf{Comparative} study (see Table \ref{tab:comparative}), we infer on \textit{sub-1006} as FINTA-m reports results only on Tractoinferno, since TrackletGPT architecture is formalized to keep generalizability in mind, it performs lesser for sub-1006 and Tractoinferno same-dataset experiments.
For an in-depth view on class-wise performance across methods (refer Figure \ref{graph}), where we compare experiments done on same and different datasets for training and testing, where we can see TrackletGPT outperforms TractoGPT\cite{goel2025tractogpt} and other methods on all inter-dataset scenarios and difficult classes (like UF, FPT etc). Here, RecoBundlesX outputs are not filtered using dMRIQC \cite{theaud2022dmriqcpy}.
TrackletGPT proves superior \textbf{generalizability} than TractoGPT in all inter-dataset scenarios (refer Table \ref{tab:inter_dataset}) and reported DICE of $0.74 \pm 0.08$ in FIESTA\cite{dumais2023fiesta} on a private dataset, Myeloinferno\cite{dumais2023fiesta}.
TrackletGPT was experimented on multiple positional embeddings, model sizes and neighbouring streamlines, where we conclude on Sinusoidal Positional Embedding, mini (9.8M), 7 neighbours, and Paragraph representation respectively.

\input{tables/inter-dataset}
\vspace{-0.2cm}
\textit{\textbf{Notation}:} "\textit{Method-Dataset}", denotes Method is trained and tested on this Dataset. Whereas, "\textit{Method-Dataset-inter}" trained on mentioned dataset and tested on the other, for example, "TrackletGPT-HCP-inter" is trained on HCP and tested on TractoInferno. With our rigorous experiments, we evaluated FINTA-m\cite{legarreta2021filtering}, RecobundlesX \cite{garyfallidis2018recognition} across the list of \textbf{23 common tracts}\cite{dumais2023fiesta} in the TractoInferno and HCP datasets. 
\input{tables/sub-1006}
\vspace{-0.2cm}
\input{tables/combined_dataset}
\vspace{-0.2cm}

\vspace{-0.2cm}
\section{Conclusion}
\vspace{-0.2cm}
In this study, we propose \textbf{TrackletGPT}, a language-like GPT framework for white matter tract segmentation, which seamlessly \textbf{generalizes} across datasets while preserving \textit{sequence} and \textit{shape} information of streamlines and bundles respectively yielding state-of-the-art inter-dataset results. We introduced \textbf{Tracklets}, which lay the groundwork for foundation models in Tractography.
\vspace{-0.2cm}
\section{Compliance with ethical standards}
\label{sec:ethics}
This research study was conducted retrospectively using human subject data made available in open access by (TractoInferno \cite{poulin2022tractoinferno}, Human Connectome Project \cite{wasserthal2018tractseg}). Ethical approval was not required as confirmed by license attached with the open access data.

\vspace{-0.4cm}
\section{Acknowledgments}
\label{sec:acknowledgments}
This work was supported by IIT Mandi from SERB CORE Research Grant with Project No: CRG/2020/005492.
\vspace{-0.4cm}



\setlength{\bibsep}{-1pt}
\renewcommand{\bibfont}{\small}
\section{References}
\vspace{-1cm}
\bibliography{refs-sq}
\end{document}

%% file: tables/inter-dataset.tex
\definecolor{pastelblue}{RGB}{173, 216, 230}   
\definecolor{pasteldarkyellow}{RGB}{240, 220, 130}   
\definecolor{pastelgreen}{RGB}{181, 234, 215}   
\begin{table}[hbt!]
    \centering
    \begin{tabular}{|p{0cm}|p{1.62cm}|p{1.4cm}|p{1.4cm}|p{1.4cm}|p{0cm}|}
    \hline
     &\textbf{Method} & \textbf{DICE} & \textbf{Overlap} & \textbf{Overreach
     }&
     \\
    \hline
    \cellcolor{pastelblue}{}t & TrackletGPT & 0.93$\pm$0.05 & 0.94$\pm$0.04 & 0.08$\pm$0.10 & \cellcolor{pastelblue}{}t \\
    \hline
    \cellcolor{pastelblue}{}t & TractoGPT & \textbf{0.96$\pm$0.04} & \textbf{0.96$\pm$0.04} & \textbf{0.04$\pm$0.06} & \cellcolor{pastelblue}{}t \\
    \noalign{\hrule height 1pt}
    \cellcolor{pasteldarkyellow}{}h & TrackletGPT & \textbf{0.90$\pm$0.07} & \textbf{0.86$\pm$0.11} & 0.05$\pm$0.06 & \cellcolor{pasteldarkyellow}{}h\\
    \hline
    \cellcolor{pasteldarkyellow}{}h & TractoGPT & 0.83$\pm$0.20 & 0.77$\pm$0.22 & \textbf{0.03$\pm$0.04} &\cellcolor{pasteldarkyellow}{}h \\
    \noalign{\hrule height 1pt}
    \cellcolor{pastelblue}{}t & TrackletGPT & \textbf{0.80$\pm$0.18} & \textbf{0.71$\pm$0.22} & \textbf{0.03$\pm$0.05} & \cellcolor{pasteldarkyellow}{}h \\
    \hline
    \cellcolor{pastelblue}{}t & TractoGPT & 0.75$\pm$0.20 & 0.67$\pm$0.28 & 0.04$\pm$0.10 & \cellcolor{pasteldarkyellow}{}h \\
    \noalign{\hrule height 1pt}
    \cellcolor{pasteldarkyellow}{}h & TrackletGPT & \textbf{0.80$\pm$0.09} & \textbf{0.81$\pm$0.15} & \textbf{0.22$\pm$0.20} & \cellcolor{pastelblue}{}t \\
    \hline
    \cellcolor{pasteldarkyellow}{}h & TractoGPT & 0.79$\pm$0.13 & 0.78$\pm$0.18 & 0.28$\pm$0.50 & \cellcolor{pastelblue}{}t \\
    \hline
    \end{tabular}
    \caption{\textbf{Inter-dataset experiments} on all combinations of Tractoinferno (blue, "t") and HCP (yellow, "h") are denoted in train column (left-most) and test column (right-most).}
    \label{tab:inter_dataset}
\end{table}

%% file: tables/sub-1006.tex
\begin{table}[hbt!]
    \centering
    \begin{tabular}{|c|c|c|c|c|c|c|}
    \hline
    \textbf{Method} & \textbf{DICE} & \textbf{Overlap} & \textbf{Overreach} \\
    \hline
    TrackletGPT & 0.94$\pm$0.08 & 0.93$\pm$0.11 & 0.04$\pm$0.09
    \\
    \hline
    RBx & 0.95$\pm$0.07 & 0.95$\pm$0.10 & 0.04$\pm$0.09 \\
    \hline
    FINTA-m & 0.79$\pm$0.13 & 0.99$\pm$0.01 & 0.59$\pm$0.59 \\
    \hline
    TractoGPT & 0.97$\pm$0.05 & 0.96$\pm$0.07 & 0.03$\pm$0.05 \\
    \hline
    \end{tabular}
    \caption{\textbf{Comparative Study}: Test results on sub-1006 Tractoinferno dataset, trained on Tractoinferno dataset}
    \label{tab:comparative}
\end{table}

%% file: tables/combined_dataset.tex
\begin{table}[hbt!]
    \centering
    \begin{tabular}{|c|c|c|c|c|c|c|}
    \hline
    \multicolumn{4}{|c|}{\textbf{Train and Test:} TractoInferno+105HCP} \\ 
    \hline
    \textbf{Representation} & \textbf{DICE} & \textbf{Overlap} & \textbf{Overreach} \\
    \hline
    Sentence & 0.87$\pm$0.08 & 0.86$\pm$0.11 & \textbf{0.14$\pm$0.23} \\
    \hline
    \textbf{Paragraph} & \textbf{0.92$\pm$0.09} & \textbf{0.93$\pm$0.07} & 0.19$\pm$0.42 \\
    \hline
    \end{tabular}
    \caption{\textbf{Ablation Study}:\textit{Average} test results across all test subjects of TractoInferno and 105HCP when trained on both TractoInferno and HCP.}
    \label{tab:ablation}
\end{table}